\documentclass[twoside,11pt]{article}

\usepackage{jmlr2e}

\usepackage{amsmath}
\usepackage{bbm} 

\usepackage{xargs}  
\usepackage[pdftex,dvipsnames]{xcolor}  

\newcommand{\fracpartial}[2]{\frac{\partial #1}{\partial  #2}}

\newcommand{\R}{\mathbb{R}}

\newcommand{\Pm}{\textnormal{P}}
\newcommand{\x}{\mathbf{x}}

\renewcommand{\mid}{\,|\,}
\newcommand{\Ts}{T^*}
\newcommand{\Cs}{C^*}

\ShortHeadings{The Brier Score under Administrative Censoring}{Kvamme and Borgan}
\firstpageno{1}

\begin{document}

\title{The Brier Score under Administrative Censoring:\\Problems and Solutions}

\author{\name{H\aa{}vard Kvamme} \email{haavakva@math.uio.no} \\
        \name{\O{}rnulf Borgan} \email{borgan@math.uio.no} \\
        \addr{Department of Mathematics\\
        University of Oslo\\
        P.O. Box 1053 Blindern\\
        0316 Oslo, Norway}
        }


\maketitle

\begin{abstract}
    The Brier score is commonly used for evaluating probability predictions.
    In survival analysis, with right-censored observations of the event times, this score can be weighted by the inverse probability of censoring (IPCW) to retain its original interpretation.
    It is common practice to estimate the censoring distribution with the Kaplan-Meier estimator, even though it assumes that the censoring distribution is independent of the  covariates.
    This paper discusses the general impact of the censoring estimates on the Brier score and shows that the estimation of the censoring distribution can be problematic.
    In particular, when the censoring times can be identified from the covariates, the IPCW score is no longer valid.
    For administratively censored data, where the potential censoring times are known for all individuals, we propose an alternative version of the Brier score.
    This administrative Brier score does not require estimation of the censoring distribution and is valid even if the censoring times can be identified from the covariates. 
\end{abstract}


\section{Introduction}
\label{sec:introduction}

Recently, there has been an increasing interest in combining machine learning methodology with survival analysis for improved time-to-event prediction.
Some methods extend the well known Cox regression with neural networks
\citep{DeepSurv, ching2018cox, yousefi2017predicting, Luck2017, Cox-Time},
while others consider a more direct approach for optimizing the likelihood for right-censored time-to-event data
\citep{biganzoli1998feed, deephit, fotso2018, gensheimer2019, article3}.
Also worth mentioning is the Random Survival Forest \citep{Ishwaran2008} which makes decision trees based on the log-rank test and estimates the cumulative hazards with the Nelson-Aalen estimator.

Although these methods are available for right-censored event times, a substantial part of the machine learning community is not familiar with survival analysis and might find it reasonable to instead apply binary classifiers for time-to-event prediction.
In short, a binary classifier estimates
the probability that an individual experience the event by time $t$, 
and can be fitted by disregarding individuals censored before that time.
As empirical evaluation of predictions is central to machine learning, the best way to convince this audience to use survival methodology would likely be to show that the survival methods give improved evaluation scores.

Arguably, the two most common evaluation criteria for survival predictions are the inverse probability of censoring weighted (IPCW) Brier score \citep{Graf1999, Gerds2006} and different versions of the concordance index \citep{Harrell1982, Antolini2005, uno2011c, gerds2013estimating}.
We will, in this paper, show that the IPCW Brier score can be biased under administrative censoring, i.e., when the right-censoring times are known for \emph{all} individuals.
In fact, we will show that this bias benefits the binary classifiers, meaning that the binary classifiers can get better scores than corresponding survival methods, even though the estimates of the binary classifiers are, in reality, much worse than those of the survival methods.
Furthermore, we will show that the IPCW Brier score is heavily dependent on the estimates of the censoring distribution, making it unattractive as an evaluation metrics as one can question the validity of the results.
We, therefore, propose the \emph{administrative Brier score}, which is created for handling both these issues.
In particular, it does not have the potential bias of the IPCW score under administrative censoring, and it does not require estimation of the censoring distribution.

To give the reader an understanding of the potential problems of the IPCW Brier score, we will start with an illustrative example in Section~\ref{sec:a_real_world_example}.
Then, in Section~\ref{sec:the_brier_score}, we will introduce the Brier score in detail and discuss more carefully the issues of the IPCW scheme.
We present our proposed alternative, the administrative Brier score, in Section~\ref{sub:a_brier_score_for_administrative_censoring}.
The binary classifiers are investigated in Section~\ref{sec:binary_classifiers_for_time_to_event_prediction}, where we will show their relationship to the potential bias of the IPCW Brier score under administrative censoring.
A simulation study is conducted in Section~\ref{sec:simulations} to empirically illustrate our findings, and in Section~\ref{sec:kkbox_churn_prediction}, we investigate a real data set with administrative censoring.
A summary and concluding remarks are made in Section~\ref{sec:discussion}.

The code for the evaluation metrics, the survival methods, the simulations, and the data sets is available at \href{https://github.com/havakv/pycox}{\texttt{github.com/havakv/pycox}}.

\section{A Real-World Example}
\label{sec:a_real_world_example}

To illustrate the potential issues with the IPCW Brier score, we consider an example encountered while researching the KKBox Churn data set in~\cite{Cox-Time}.
The task is to predict whether or not customers continue to subscribe to the KKBox music streaming service $t$ days after their first subscription.
If customers leave their subscription they have \emph{churned}, and these are the events we want to model.
The follow-up times of the customers depend on when they first subscribed, as we only follow customers up to January~29, 2017.
A substantial part of the customers do not experience the event before this date and are instead \emph{right-censored}.
The type of right-censoring encountered here is called \emph{administrative censoring} or \emph{progressive type I censoring}, meaning that we actually know the censoring times of \emph{all} the customers.
So even for customers for whom we have observed the churn time, we know the time they would have been censored. 

We approach the modeling of the event-time distribution in two ways.
The first is with the Logistic-Hazard method \citep{brown1975use, biganzoli1998feed, gensheimer2019, article3} which accounts for censored observations by considering  the likelihood for right-censored event times.
We use the version of the method described by \citet{article3}, meaning that we parameterize the discrete hazards with a neural network in the form of a multilayer perceptron (MLP).

The second approach is to fit a binary classifier for each time $t$ and remove all customers censored before this time.
The responses (labels) given to the classifiers are indicators of whether or not each customer has churned.
We denote this as the \emph{BCE} method, as it minimizes the binary cross-entropy of the survival estimates, where survival means that a customer has not yet churned.
The BCE method is an MLP with equivalent network structure to that of the Logistic-Hazard, with each output node corresponding to a binary classifier at time $t$.
The method will be described in detail in Section~\ref{sec:binary_classifiers_for_time_to_event_prediction}.
As censored individuals are removed from the data set at their time of censoring, the BCE method has a bias towards higher churn probabilities. 
We would, therefore, expect the Logistic-Hazard to perform better for the KKBox data set.

\begin{figure}[t]
    \centering
    \includegraphics[width=.99\linewidth]{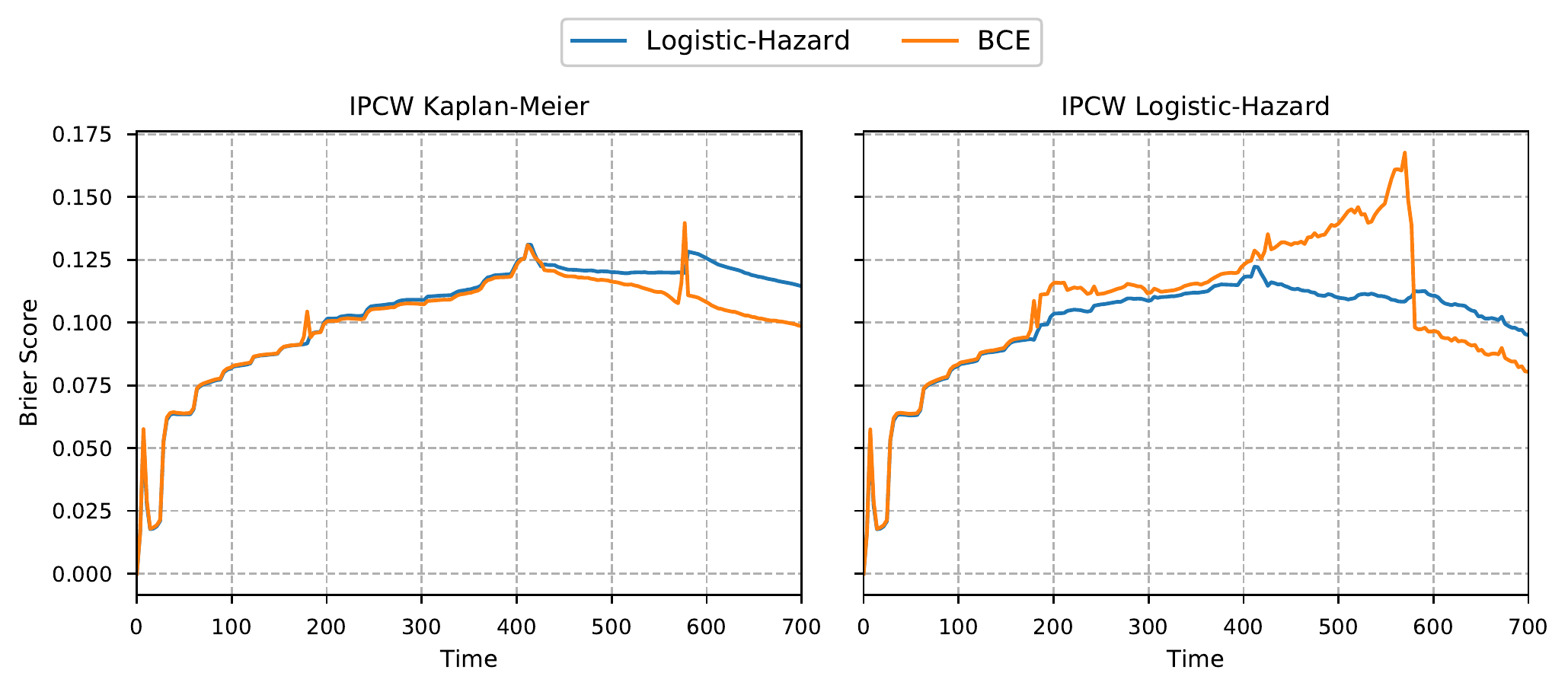}
    \vspace{-0.3cm}
    \caption{Brier scores of survival estimates from the Logistic-Hazard method and the BCE method on the KKBox data set. The Brier scores are weighted by the inverse probability of censoring estimated with Kaplan-Meier (left) and with a Logistic-Hazard model (right).}\label{fig:kkbox_bs_km}
\end{figure}

The IPCW Brier scores with Kaplan-Meier censoring estimates \citep{Graf1999} are calculated for the survival estimates of the two methods, and a plot of the results is shown in the left part of Figure~\ref{fig:kkbox_bs_km}.
For higher times, we see that the BCE method has better scores than the Logistic-Hazard method.
This is unexpected, as the censoring proportion increases with time, meaning the bias of the BCE estimates increases with time.
However, the Kaplan-Meier censoring estimates are not covariate dependent, so the results are likely explained by poor estimation of  the censoring distribution.
This was addressed by \citet{Gerds2006}, but the Kaplan-Meier estimator continues to be the most common choice for estimating the weights in the IPCW Brier score.

In the right part of Figure~\ref{fig:kkbox_bs_km}, we have plotted the IPCW Brier scores where the censoring distribution is estimated with a Logistic-Hazard model similar to that of the survival estimates.
We now see that the BCE generally performs worse than the Logistic-Hazard, but we still
find that it gets better scores than the Logistic-Hazard for the largest times.
We do, however, believe that the BCE actually performs worse than the Logistic-Hazard, also for the largest times, and that the reason for these results can be explained by some artifacts of the IPCW Brier scores applied to a data set with administrative censoring.

In the following sections, we will present what we believe to be the cause of these results, and we will use simulated data to reproduce the artifacts.
We will, also, propose the \emph{administrative Brier score}, which does not suffer from the same vulnerabilities as the IPCW Brier score, and does not even require estimation of the censoring distribution.
In Section~\ref{sec:kkbox_churn_prediction}, we will revisit the KKBox data set for a more in-depth analysis.

\section{The Brier Score}
\label{sec:the_brier_score}

In the following, we present the Brier score for evaluating time-to-event predictions in the form of survival estimates.
We will then introduce the topic of right-censoring in survival analysis, and show how the IPCW Brier score accounts for censored observations.
This is followed by a discussion of administratively censored observations and how they affect the IPCW Brier score.
We end the section by introducing a new Brier score for handling administratively right-censored observations.

Let $\Ts_i$ denote the event time of individual $i$ and let $f_i(t)$ denote the density function of this event time.
The survival function of individual $i$ is then given by
\begin{align*}
    S_i(t) = \Pm(\Ts_i > t) = \int_t^\infty f_i(u)\, du. 
\end{align*}
In time-to-event prediction, we want to estimate $S_i(t)$ for all individuals, and we will let $\pi_i(t)$ denote these estimates.
To simplify the presentation, we will disregard the estimation uncertainty, and consider the $\pi_i(t)$'s as known (non-random) functions.

A reasonable metric for evaluating the predictive performance of the $\pi_i(t)$'s, is to calculate the mean squared error of the estimates to the true survival functions.
For a data set with $n$ individuals, this score is
\begin{align}
    \label{eq:mse}
    \text{MSE}(t) = \frac{1}{n} \sum^{n}_{i=1} {\left[S_i(t) - \pi_i(t) \right]}^2.
\end{align}
However, $S_i(t)$ is not known outside of simulations, and we instead observe event times $\Ts_i$ drawn from the event time distribution.
The Brier score for uncensored data approximates the true survival functions with step-functions with jumps at the event times, giving
\begin{align}
    \label{eq:full_bs}
    \text{BS}(t)
    &= \frac{1}{n} \sum_{i=1}^n {\left[ \mathbbm{1}\{\Ts_i > t\} - \pi_i(t) \right]}^2 \\
    &= \frac{1}{n} \sum^{n}_{i=1} \left[ {\pi_i(t)}^2 \mathbbm{1}\{T_i^* \leq t\} + {[1 - \pi_i(t)]}^2 \mathbbm{1}\{T_i^* > t\} \right ] \nonumber
\end{align}
The expectation of the Brier score is
\begin{align*}
    \mathbb{E} \left[\text{BS}(t) \right]
    &= \frac{1}{n} \sum^{n}_{i=1} \mathbb E \left[ {\pi_i(t)}^2 \mathbbm{1}\{T_i^* \leq t\} + {[1 - \pi_i(t)]}^2 \mathbbm{1}\{T_i^* > t\} \right ]\\
    &= \frac{1}{n} \sum^{n}_{i=1} \left[ {\pi_i(t)}^2\, \Pm(T_i^* \leq t) + {[1 - \pi_i(t)]}^2\, \Pm(T_i^* > t) \right] \\
    &= \frac{1}{n} \sum^{n}_{i=1} \left[ {\pi_i(t)}^2\, [1 - S_i(t)] + {[1 - \pi_i(t)]}^2\, S_i(t) \right] \\
    &= \frac{1}{n} \sum^{n}_{i=1} \left( {\left[ S_i(t) - \pi_i(t)\right]}^2 + S_i(t) \left[1 - S_i(t) \right] \right) \\
    &= \text{MSE}(t) + \frac{1}{n} \sum^{n}_{i=1} S_i(t) \left[1 - S_i(t) \right].
\end{align*}
So the expected Brier score is the sum of the MSE in~\eqref{eq:mse} and a constant that does not depend on our survival estimates $\pi_i(t)$. 
This constant is the irreducible error of approximating the true survival functions $S_i(t)$ with the step-functions $\mathbbm{1}\{\Ts_i > t\}$.
So minimizing the expected Brier score is equivalent to minimizing the MSE\@, and the minimum is obtained for the true survival functions, i.e., $\pi_i(t) = S_i(t)$.

\subsection{The IPCW Brier Score}
\label{sub:the_ipcw_brier_score}

In many applications, only a subset of the event times $\Ts_i$ is observed.
For some individuals, we instead only know that the event time occurs after some censoring time $\Cs_i$.
For modeling these data sets, we consider the right-censored event time $T_i = \min\{\Ts_i,\, \Cs_i\}$ and the event indicator $D_i = \mathbbm{1}\{\Ts_i \leq \Cs_i\}$.
We follow the common convention in survival analysis that when an event and censoring time coincide, we observe the occurrence of the event.

As we only have partial information, the Brier score in~\eqref{eq:full_bs} cannot be calculated.
We can, however, approximate it by weighting the scores of the observed event times by the inverse probability of censoring \citep{Graf1999, Gerds2006}.
This is called \emph{inverse probability of censoring weighting} (IPCW), and the IPCW Brier score is given by
\begin{align}
    \label{eq:bs_ipcw_n}
    \text{BS}_\text{IPCW}(t) = \frac{1}{n} \sum_{i=1}^n \left [
        \frac{\pi_i{(t)}^2 \mathbbm{1}\{T_i \leq t, D_i = 1\}}{G_i(T_i-)} +
        \frac{{[1 - \pi_i(t)]}^2 \mathbbm{1}\{T_i > t\}}{G_i(t)} \right ],
\end{align}
where $G_i(t) = \Pm(\Cs_i > t) > 0$, is the survival function of the censoring distribution for individual $i$.
Assuming, for simplicity, that the $G_i(t)$'s are known functions, the expectation of the IPCW Brier score is
\begin{align*}
    \mathbb{E}\left[ \text{BS}_\text{IPCW}(t) \right]
    &= \frac{1}{n} \sum^{n}_{i=1} \mathbb E \left[ \frac{{\pi_i(t)}^2 \mathbbm{1}\{T_i \leq t, D_i=1\}}{G_i(T_i-)} +
    \frac{{[1 - \pi_i(t)]}^2 \mathbbm{1}\{T_i > t\}} {G_i(t)} \right]\\
    &=  \frac{1}{n} \sum^{n}_{i=1} {\pi_i(t)}^2\, \mathbb E \left[\frac{\mathbbm{1}\{T_i^* \leq t, T_i^* \leq C_i^*\}}{G_i(T_i^*-)} \right] +
    {[1 - \pi_i(t)]}^2\, \frac{\Pm(T_i^* > t, C_i^* > t)}{G_i(t)}\\
    &=  \frac{1}{n} \sum^{n}_{i=1} {\pi_i(t)}^2\, \int_0^t \frac{G_i(u-)\, f_i(u)}{G_i(u-)}\, du  +
    {[1 - \pi_i(t)]}^2\, \frac{G_i(t)\, S_i(t)}{G_i(t)}\\
    &=  \frac{1}{n} \sum^{n}_{i=1} {\pi_i(t)}^2\, [1 - S_i(t)] + {[1 - \pi_i(t)]}^2\, S_i(t),\\
    &= \text{MSE}(t) + \frac{1}{n} \sum^{n}_{i=1} S_i(t) \left[1 - S_i(t) \right].
\end{align*}
We see that this is identical to the expectation of the uncensored Brier score~\eqref{eq:full_bs}, so the IPCW Brier score is a reasonable approximation of the uncensored Brier Score.

\subsubsection{IPCW with Administrative Censoring}
\label{sub:ipcw_with_administrative_censoring}

For administrative censoring, \emph{all} the censoring times $\Cs_i$ are known, regardless of whether individual $i$ experienced an event or was censored.
Although $\Cs_i$, in this case, is not random, we will continue to capitalize $\Cs_i$ for simplicity.
As we know the censoring times, the censoring distribution is given by the step-function $G_i(t) = \mathbbm{1}\{\Cs_i > t\}$, and the IPCW Brier score~\eqref{eq:bs_ipcw_n} can be written as
\begin{align}
    \label{eq:ipcw_admin_n}
    \text{BS}_\text{IPCW}(t) = \frac{1}{n} \sum_{i=1}^n \left[
        {\pi_i(t)}^2 \mathbbm{1}\{T_i \leq t, D_i = 1\} +
        {[1 - \pi_i(t)]}^2 \mathbbm{1}\{T_i > t\} \right].
\end{align}
This is equivalent to the unweighted Brier score~\eqref{eq:full_bs} on the subset of individuals that are not censored,
meaning we simply disregard the set of censored individuals 
$\{i:\, T_i \leq t, D_i  =  0\}$.
The step-function $G_i(t) = \mathbbm{1}\{\Cs_i > t\}$ violates the assumptions of the IPCW Brier score which, as stated in Section~\ref{sub:the_ipcw_brier_score}, requires $G_i(t) > 0$.
As a result, the expectation of the score is no longer equal to that of the uncensored Brier score,
\begin{align*}
    \mathbb{E}\left[ \text{BS}_\text{IPCW}(t) \right]
    &= \frac{1}{n} \sum^{n}_{i=1} \mathbb E \left[ {\pi_i(t)}^2 \mathbbm{1}\{T_i \leq t, D_i = 1\} + {[1 - \pi_i(t)]}^2 \mathbbm{1}\{T_i > t\} \right]\\
    &= \frac{1}{n} \sum^{n}_{i=1} \left[ {\pi_i(t)}^2\, \Pm(T_i \leq t, D_i = 1) + {[1 - \pi_i(t)]}^2\, \Pm(T_i > t) \right] \\
    &= \frac{1}{n} \sum^{n}_{i=1} \left[  {\pi_i(t)}^2\, \Pm(T_i^* \leq \min\{t, C_i^*\}) + {[1 - \pi_i(t)]}^2\, \Pm(T_i^* > t) \mathbbm{1}\{C_i^* > t\} \right]
\end{align*}

\begin{align*}
    \quad\quad\quad &= \frac{1}{n} \sum^{n}_{i=1} \left[  {\pi_i(t)}^2\, \left[1 - S_i(\min\{t, C_i^*\})\right] + {[1 - \pi_i(t)]}^2\, S_i(t) \mathbbm{1}\{C_i^* > t\} \right]\\
    &= \frac{1}{n} \sum^{n}_{i=1} \left[ {\pi_i(t)}^2\, [1 - S_i(t)] + {[1 - \pi_i(t)]}^2\, S_i(t) \right] \mathbbm{1}\{\Cs_i > t\}\\
    &\quad + \frac{1}{n} \sum^{n}_{i=1} {\pi_i(t)}^2\, [1 - S_i(C_i^*)]\,  \mathbbm{1}\{\Cs_i \leq t\}.
\end{align*}
We see that the expectation is only equal to that of the uncensored Brier score if all individuals have censoring times $\Cs_i > t$.
For this group, with $\Cs_i > t$, we still have that the minimizers of the score are $\pi_i(t) = S_i(t)$.
However, for the other group, with $\Cs_i \leq t$, we see that the minimizers are $\pi_i(t) = 0$.
Hence, according to this score, the optimal survival estimates are $\pi_i(t) = S_i(t)\, \mathbbm{1}\{\Cs_i > t\}$.
This is problematic, as we want an evaluation metric that decreases as our estimates approach the true survival functions $S_i(t)$.

Due to these biases, one would, in practice, not use the Brier score~\eqref{eq:ipcw_admin_n} with $G_i(t) = \mathbbm{1}\{\Cs_i > t\}$, but instead use the regular IPCW Brier score~\eqref{eq:bs_ipcw_n} with estimates of the censoring distributions.
We will, however, see that these issues can still be present.
Specifically, if the covariates hold sufficient information to identify the administrative censoring times, then the estimated censoring distributions can approach the step functions $G_i(t) = \mathbbm{1}\{\Cs_i > t\}$, and we get a Brier score close to that of~\eqref{eq:ipcw_admin_n}.

\subsubsection{Estimation of the Censoring Distribution}
\label{sub:estimating_the_censoring_distribution}

In practice, we need to estimate the censoring distributions $G_i(t)$ to use the IPCW Brier score.
This estimation is, in itself, a time-to-event prediction problem, and can be addressed in the same manner as the original time-to-event problem.
\citet{Graf1999} proposed to use the Kaplan-Meier estimates of the censoring distribution, and this is still the most common approach.
However, the Kaplan-Meier estimator disregards the covariates, meaning all individuals are assumed to have the same censoring distribution.
This can lead to biased censoring estimates, as addressed by \citet{Gerds2006}.

In predictive modeling, we typically split our data set into
a training set used to fit the  models,
a validation set used for hyperparameter tuning,
and a test set used for evaluating the models' predictions.
We, therefore, only consider the Brier score calculated on a test set in this paper.
Both \citet{Graf1999} and \citet{Gerds2006} use the test set to estimate the censoring distribution, which is reasonable for simple models such as the Kaplan-Meier estimator and Cox regression.
If we, however, want to use more flexible models, such as the Logistic-Hazard with neural networks, fitting to the test set is likely to results in  overfitted censoring estimates.
To the best of our knowledge, this topic has not been addressed in the literature, so there are no ``best practices'' for how to approach such estimation problems.
In this paper, we will, therefore, treat the censoring distribution in the same manner as the event-time distribution, meaning we fit the censoring model to the training set, and use a validation set to verify that the estimates are reasonable.

When the censoring distribution is obtained from the Kaplan-Meier estimator on the test set, the weights of the IPCW Brier score~\eqref{eq:bs_ipcw_n} sum to $n$.
But for more flexible methods, and methods fitted to the training set, this is not necessarily the case.
A more general version of the IPCW Brier score~\eqref{eq:bs_ipcw_n}, that works for any reasonable censoring estimates, is obtained by replacing $n$ with the sum of the weights 
\begin{align*}
    \tilde n(t) = \sum_{i=1}^n \left [
        \frac{\mathbbm{1}\{T_i \leq t, D_i = 1\}}{G_i(T_i-)} +
        \frac{\mathbbm{1}\{T_i > t\}}{G_i(t)} \right ].
\end{align*}
This ensures that the Brier score is between 0 and 1,
\begin{align}
    \label{eq:bs_ipcw}
    \text{BS}_\text{IPCW}(t) = \frac{1}{\tilde n(t)} \sum_{i=1}^n \left [
        \frac{\pi_i{(t)}^2 \mathbbm{1}\{T_i \leq t, D_i = 1\}}{G_i(T_i-)} +
        \frac{{[1 - \pi_i(t)]}^2 \mathbbm{1}\{T_i > t\}}{G_i(t)} \right ].
\end{align}
This is the version of the IPCW Brier score we use in all our experiments.

The censoring estimates in~\eqref{eq:bs_ipcw} are actually functions of the covariates, $G_i(t) = G(t \mid \x_i)$.
If there is enough information in the covariates $\x_i$ to identify the censoring times $\Cs_i$, the censoring estimates will approach the step-functions $G_i(t) = \mathbbm{1}\{\Cs_i > t\}$, meaning the scores approach the biased Brier score~\eqref{eq:ipcw_admin_n}.
As an example of this, consider a study where all censoring is due to administrative censoring at the closure of the study, and one includes the start date for each individual as a covariate, then the $\Cs_i$'s can be identified.
A more realistic example would be that a combination of certain covariates can identify the start date of some individuals and, consequently, a subset of the censoring times can be identified. 

If we estimate the censoring distribution with a flexible method, we might experience that some of the estimates of $G_{i}(t)$ become very small.
This corresponds to very large weights, meaning that a single individual can potentially dominate the score.
To prevent this from occurring, we set a maximum weight allowed.
As an example, if we have a maximum weight of 100, we do not allow estimates of $G_{i}(t) < 0.01$, giving the interpretation that  a single individual can maximally represent 100 individuals.
In practice, this is ensured by setting weights larger than 100 to be 100.
By decreasing the maximum allowed weight, we reduce the variance of the IPCW Brier score at the expense of introducing some bias.

It should now be clear that the IPCW Brier score's dependence on the censoring estimates is undesirable, as we do not want an evaluation criterion to be heavily dependent on the choices made by the researcher.
If the scores are substantially affected by the censoring estimates, one can question the validity of the results.

\subsection{The Administrative Brier Score}
\label{sub:a_brier_score_for_administrative_censoring}

We have shown that the IPCW Brier score may have undesirable behavior under administrative censoring, and that estimation of the censoring distribution can be problematic in general.
We propose an alternative that alleviates both of these problems.
Our approach is to use the uncensored Brier score~\eqref{eq:full_bs} and simply remove individuals from evaluation after their administrative censoring time. This is possible as we know the censoring times $\Cs_i$ for all individuals.
However, this also means that the score is not applicable in studies where we do not know all the censoring times.

To define the \textit{administrative Brier score}, first note that for $\Cs_i \geq t$ we know whether $\Ts_i \leq t$ or $\Ts_i > t$ (remember that if $\Ts_i = \Cs_i$ we assume that we observe the occurrence of the event).
The administrative Brier score is then
\begin{align}
    \label{eq:bs_admin}
    \text{BS}_\text{A}(t)
    &= \frac{1}{\tilde{n}(t)} \sum_{i=1}^n {\left[\mathbbm{1}\{\Ts_i > t\}  - \pi_i(t) \right]}^2\, \mathbbm{1}\{\Cs_i \geq t\},\nonumber\\
    &= \frac{1}{\tilde{n}(t)} \sum_{i:\, \Cs_i \geq t} {\left[\mathbbm{1}\{\Ts_i > t\}  - \pi_i(t) \right]}^2,
\end{align}
where we scale by
\begin{align*}
    \tilde{n}(t) = \sum^{n}_{i=1} \mathbbm{1}\{\Cs_i \geq t\},
\end{align*}
instead of $n$.
The expectation of the administrative Brier score is
\begin{align*}
    \mathbb{E}\left[ \text{BS}_\text{A}(t) \right]
    &= \frac{1}{\tilde{n}(t)} \sum_{i=1}^n \mathbb{E} \left[ {\left[\mathbbm{1}\{\Ts_i > t\}  - \pi_i(t) \right]}^2 \right]\, \mathbbm{1}\{\Cs_i \geq t\} \\
    &= \frac{1}{\tilde{n}(t)} \sum^{n}_{i=1}  \left[{\pi_i(t)}^2\, [1 - S_i(t)] + {[1 - \pi_i(t)]}^2\, S_i(t) \right] \mathbbm{1}\{\Cs_i \geq t\}\\
    &= \frac{1}{\tilde{n}(t)} \sum_{i:\, \Cs_i \geq t}  \left[{\pi_i(t)}^2\, [1 - S_i(t)] + {[1 - \pi_i(t)]}^2\, S_i(t) \right],
\end{align*}
which, for the subset of individuals with $\Cs_i \geq t$, is identical to the uncensored Brier score.
Individuals with $\Cs_i < t$ give no contribution to the score.
As we have no information about event times $\Ts_i > \Cs_i$, it is reasonable to only consider this subset.

In practice, it is probably reasonable that a subset of the administrative censoring times can be identified by the covariates. 
We saw in Sections~\ref{sub:ipcw_with_administrative_censoring} and~\ref{sub:estimating_the_censoring_distribution} that the IPCW Brier score can be biased for this type of data.
As the administrative Brier score is still minimized by the true survival functions $\pi_i(t) = S_i(t)$, it is more robust to these issues than the IPCW Brier score.

\section{Binary Classifiers for Time-to-Event Prediction}
\label{sec:binary_classifiers_for_time_to_event_prediction}

In machine learning, it is quite common to approach time-to-event prediction with binary classifiers rather than relying on survival methodology.
Especially for someone without experience in survival analysis, a binary classifier may seem like a reasonable approach to time-to-event modeling. 

A binary classifier can be constructed for a given time $t$ by minimizing the binary cross-entropy (negative log-likelihood for Bernoulli data) and disregarding individuals that were censored before time $t$. 
This gives the loss function,
\begin{align}
    \label{eq:bce_loss_t}
    \text{loss}_\text{BCE}(t) 
    &= - \sum^{n}_{i=1} \Big(\mathbbm{1}\{T_i > t\} \log[\pi_i(t)] + \mathbbm{1}\{T_i \leq t,\, D_i = 1\} \log[1 - \pi_i(t)] \Big)\\
    &= - \sum^{n}_{i=1} \Big(y_i \log[\pi_i(t)] + (1-y_i) \log[1 - \pi_i(t)] \Big)
    \, \Big(1 - \mathbbm{1}\{T_i \leq t,\, D_i=0 \}\Big). \nonumber
\end{align}
were the labels $y_i = \mathbbm{1}\{T_i > t\}$ denote if the events happen after time $t$.
If we want survival estimates for a range of times $\tau_1 < \tau_2 < \cdots < \tau_m$, we can fit a model for each 
$\tau_j$.
Alternatively, if we use a model for $\pi_i(\tau_j)$ that can be estimated for multiple $\tau_j$'s simultaneously, such as a neural network with $m$ output nodes, we can fit the model to the sum of the $m$ losses
\begin{align}
    \label{eq:bce_loss_full}
    \text{loss}_\text{BCE} = -  \sum_{j=1}^m \sum^{n}_{i=1}
    \Big(\mathbbm{1}\{T_i > \tau_j\} \log[\pi_i(\tau_j)] + \mathbbm{1}\{T_i \leq \tau_j,\, D_i = 1\} \log[1 - \pi_i(\tau_j)] \Big).
\end{align}
We refer to this approach as the BCE method.

The binary classifiers and BCE method are clearly biased, as the removal of censored individuals decreases the survival estimates.
In fact, if there is sufficient information in the covariates to identify the censoring times $\Cs_i$,
the survival estimates of the binary classifiers will approach
\begin{align}
    \label{eq:bce_optimal_pi}
    \pi_i(t) = S_i(t) \mathbbm{1}\{ \Cs_i > t\}.
\end{align}
The derivations leading to~\eqref{eq:bce_optimal_pi} are given in Appendix~\ref{sec:the_bce_survival_estimates}.
We recognize these estimates as the minimizers of the IPCW Brier score in Section~\ref{sub:ipcw_with_administrative_censoring}.
So given that the $\Cs_i$'s can be identified from the covariates and we have a sufficiently large data set, the binary classifiers are essentially optimal for minimizing the IPCW Brier score.
If the covariates only identify a subset of the $\Cs_i$'s, it is not clear whether or not a binary classifier will give better IPCW Brier scores than a corresponding survival method.
We believe this might explain why the BCE method got a lower Brier score on the KKBox data set in Section~\ref{sec:a_real_world_example}.

If a researcher ignores censored individuals in the test set in the same manner as for the binary classifiers~\eqref{eq:bce_loss_t}, 
the uncensored Brier score~\eqref{eq:full_bs} will take the form of the IPCW Brier score in~\eqref{eq:ipcw_admin_n} with $G_i(t) = \mathbbm{1}\{\Cs_i > t\}$,
\begin{align*}
    \text{BS}(t) = \frac{1}{n} \sum_{i=1}^n \left[
        {\pi_i(t)}^2 \mathbbm{1}\{T_i \leq t, D_i = 1\} +
        {[1 - \pi_i(t)]}^2 \mathbbm{1}\{T_i > t\} \right].
\end{align*}
We know from Section~\ref{sub:ipcw_with_administrative_censoring} that the expectation of this Brier score is minimized by the estimates~\eqref{eq:bce_optimal_pi}.

\subsection{Simulation with BCE}
\label{sub:simulation_with_bce}

To illustrate the concerns addressed in the previous sections, we conduct a simple simulation study.
We draw event times from a discrete event-time distribution with 1000 discrete time points and constant hazard $h_i(t) = 0.00084$.
The censoring times are drawn uniformly over the full duration span from 0 to 100.
We will use the censoring times in two distinct ways.
The first is to consider the censoring times as random, which corresponds to a linear survival function $G_i(t)$ for the censoring, identical for all individuals.
The second approach is to consider the censoring times identifiable from the covariates, meaning we have step-functions $G_i(t) = \mathbbm{1}\{\Cs_i > t\}$.
In this case, we will use a monotone function of $\Cs_i$ as a covariate, which represents the inclusion of covariates that can be used to identify the censoring time.
For brevity, we will denote this covariate as $\Cs_i$, as it is equivalent to using $\Cs_i$ as a covariate.
In summary, the only distinction between the two approaches is whether or not the covariates contain information about the censoring time $\Cs_i$.

\begin{figure}[t]
    \centering
    \includegraphics[width=1.0\linewidth]{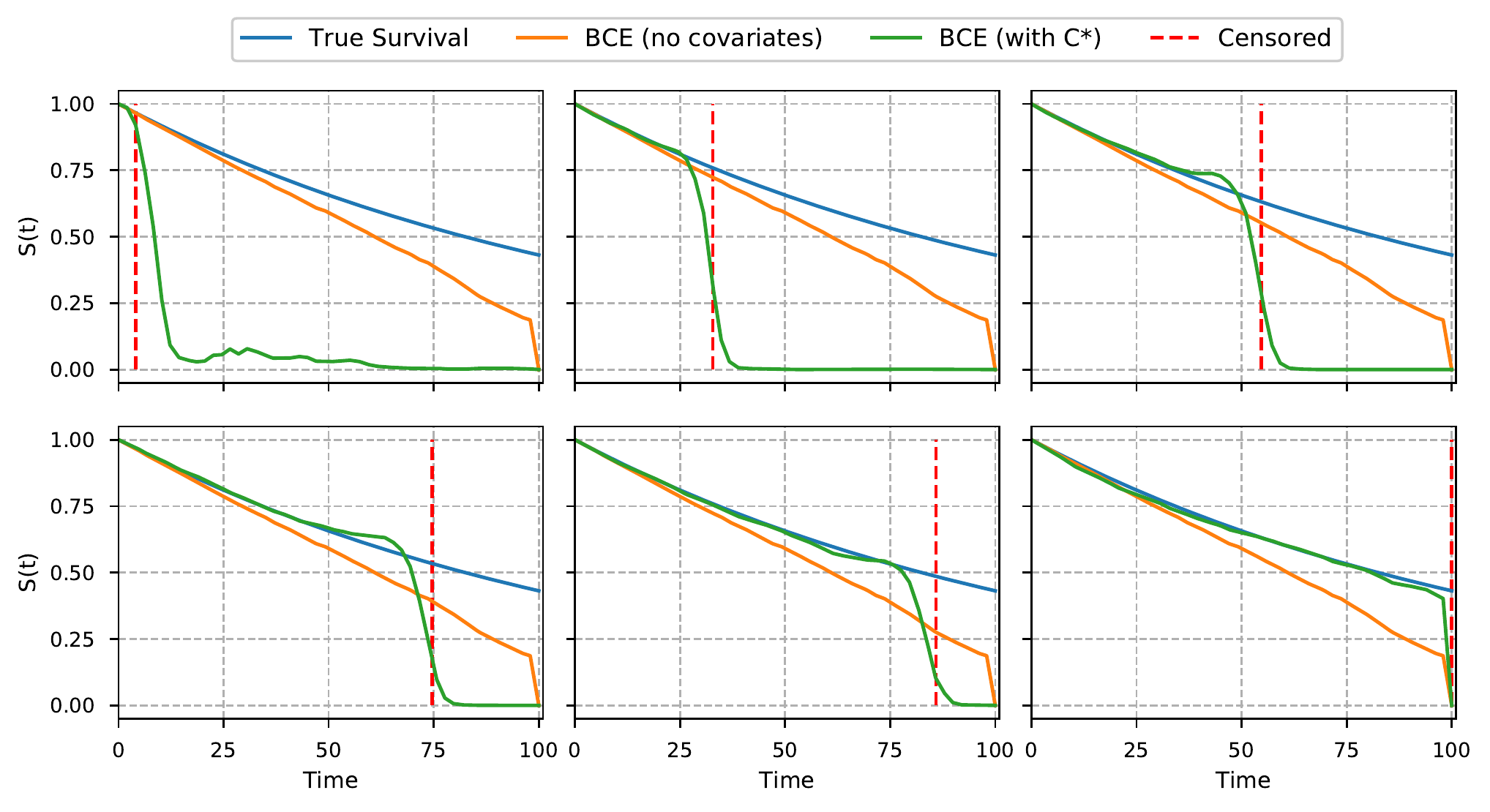}
    \vspace{-0.6cm}
    \caption{Survival estimates from the BCE method with and without a monotone function of the censoring time $\Cs_i$ as a covariate. The vertical dotted red line gives the censoring time $\Cs_i$. The true survival function is plotted in blue. All event times are drawn from the same distribution.}\label{fig:surv_bce}
\end{figure}

We fit two neural networks with the BCE loss~\eqref{eq:bce_loss_full} to 10,000 simulated samples, the first without the covariate identifying $\Cs_i$ and the second with this covariate.
In Figure~\ref{fig:surv_bce} we have plotted the estimated survival function for six distinct censoring times $\Cs_i$, marked by the vertical red dotted lines.
The plots also contain the true survival function (in blue) for comparison.
From the orange lines, we see a clear bias of the BCE method (binary classifier) applied to right-censored data, as it always underestimates the survival.
On the other hand, the survival estimates by the BCE method with censoring information, represented by the green lines, follow the true survival function reasonably well up till the censoring times and then fall to zero.
This agrees well with the optimal estimates in~\eqref{eq:bce_optimal_pi}.

\begin{figure}[t]
    \centering
    \includegraphics[width=1.0\linewidth]{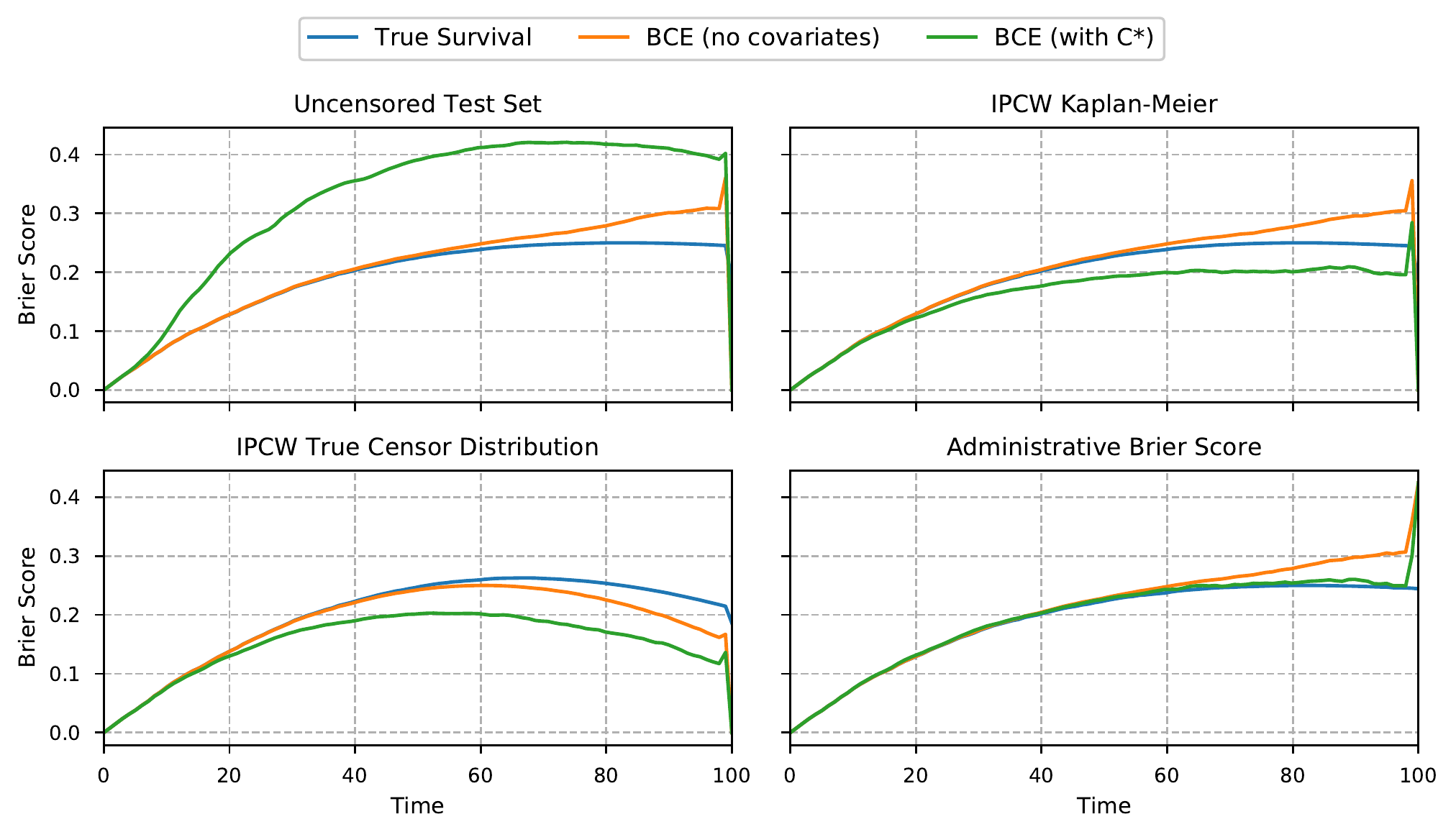}
    \vspace{-0.6cm}
    \caption{Brier scores from simulations with administrative censoring. The top left plot uses an uncensored test set, while the three other use the right-censored test set. The orange lines represent a BCE method without any information about the censoring times, while the green lines represent a BCE method with sufficient covariate information to identify the censoring times.}\label{fig:bs_bce}
\end{figure}

Next, we investigate the resulting Brier scores of these survival estimates.
In Figure~\ref{fig:bs_bce} we have plotted the uncensored Brier score (top left), the IPCW Brier scores~\eqref{eq:bs_ipcw} with Kaplan-Meier estimates of $G_i(t)$ (top right), the IPCW scores with the true censoring distribution $G_i(t) = \mathbbm{1}\{\Cs_i > t\}$ (bottom left), and the proposed administrative Brier score (bottom right) given by~\eqref{eq:bs_admin}.
Recall that the IPCW scores with the true censoring distribution (bottom left) correspond to removing censored individuals in the same manner as that of a binary classifier.
The uncensored Brier score is computed on an uncensored test set, while the three other scores are computed on a censored test set with the same censoring distributions as in the training set.

From the uncensored Brier score, it is clear that the true survival function does better than the two BCE methods.
However, for both the IPCW Brier scores we see that the BCE method with censoring information performs the best.
In fact, when we use the true censoring step-function as weights, even the BCE without knowledge of the censoring time has better scores than the true survival function.
These figures, therefore, clearly illustrate a potential weakness of the IPCW applied to data sets with identifiable censoring times.

The administrative Brier score (bottom right), rightfully identifies the true survival function as better than the estimates of the BCE methods.
For this score, the BCE with censoring information performs very closely to the true survival function, but this is expected as the survival estimates $\pi_i(t)$ are close to the true survival function for $\Cs_i > t$.
 
The simulations were made to represent a worst-case scenario for the IPCW Brier score, in the sense  that the covariates contain information that makes the censoring times $\Cs_i$ easily identifiable for all individuals.
It is unlikely to find such extreme results in real data sets, but it serves the purpose of illustrating the potential issues.

\section{Simulations}
\label{sec:simulations}

The simulations in Section~\ref{sub:simulation_with_bce} illustrated the potential issues of using the IPCW Brier score on a data set with administrative censoring.
The simulations were, however, tailored to emphasize such issues by having a very simple survival function, and censoring times $\Cs_i$ that were very easy to identify from the covariates.
In reality, the scores depend on how well the BCE method is able to estimate both the survival and censoring distribution.
Hence, we conduct a more reasonable simulation study to investigate the extent of these issues.

We will use the framework presented by \citet{article3} to create simulated data sets.
This means that we draw event times by sequentially sampling from discrete-time hazards on the time grid
$\{0.1, 0.2,\ldots,100\}$.
The hazards are specified through the logit-hazard function $g(t \mid \x) \in \R$.
Note that $g(t \mid \x)$ is just the notation used by \citet{article3}, and is not related to $G_i(t)$.
The discrete-time hazards are given by the sigmoid
\begin{align*}
    h(t \mid \x) = \frac{1}{1 + \exp\left[-g(t \mid \x) \right]}.
\end{align*}
The logit-hazards $g(t \mid \x)$ are made up of a weighted sum of three functions $g_\text{sin}(t \mid \x)$, $g_\text{con}(t \mid \x)$,  and $g_\text{acc}(t \mid \x)$,
\begin{align}
    \label{eq:g_all}
    g(t \mid \x) &= \alpha_1\, g_\text{sin}(t \mid \x) + \alpha_2\, g_\text{con}(t \mid \x) + \alpha_3\, g_\text{acc}(t \mid \x), \\
    g_\text{sin}(t \mid \x)  &= \gamma_1 \sin(\gamma_2 [ t + \gamma_3]) + \gamma_4, \nonumber\\
    \label{eq:g_con}
    g_\text{con}(t \mid \x)  &= \gamma_5, \\
    g_\text{acc}(t \mid \x)  &= \gamma_6 \cdot t - 10, \nonumber\\
    \alpha_i &= \frac{\exp(\gamma_{i+6})}{\sum_{j=1}^3 \exp(\gamma_{j+6})}, \quad \text{for } i=1, 2, 3. \nonumber
\end{align}
Here each of the nine $\gamma_k$'s is a function of the covariates $\x$, meaning $\gamma_k$ is just a simplified notation of $\gamma_k(\x)$.
The exact definitions of these $\gamma_k(\x)$'s are given in \citet[Appendix~A.1]{article3} which also explains the scheme used to draw the covariates.
With $\tau_j$ denoting the $j$'th time point in $\{0.1, 0.2, \ldots, 100\}$, the survival function is given by
\begin{align}
    \label{eq:sim_theo_surv}
    S_i(\tau_j) = S(\tau_j \mid \x_i) = \prod_{k=1}^j \left[1 - h(\tau_k \mid \x_i) \right].
\end{align}
To incorporate administrative censoring in the simulations, we consider a monotonically decreasing function $Q(t \mid \x_i)$, and let the censoring time $\Cs_i$ be defined by a threshold $\epsilon$ such that $Q(\Cs_i \mid \x_i) = \epsilon$.
This gives the survival function of the censoring distribution
\begin{align}
    \label{eq:sim_censor_surv}
    G_i(t) = \mathbbm{1}\{Q(t \mid \x_i) > \epsilon\}
\end{align}
Hence, the censoring is deterministic, while the complexity of $Q(t \mid \x_i)$ controls the complexity of the relationship between the covariates $\x_i$ and the  censoring time $\Cs_i$.
We will let $Q(t \mid \x_i)$ have the same functional form as a survival function, meaning it is defined in the same manner as~\eqref{eq:sim_theo_surv}, but with its own set of covariates independent of the covariates of the event-time distribution.
In all the experiments we set $\epsilon = 0.2$.

In the experiments, we compare the BCE method with the Logistic-Hazard method.
We use the  Logistic-Hazard because of its similarity to the BCE method, in that it also minimizes the binary cross-entropy, but use the discrete hazards instead of the survival estimates.
We use the implementation of the Logistic-Hazard by \citet{article3}, but the method has been described by multiple authors \citep{brown1975use, biganzoli1998feed, gensheimer2019}.

\subsection{Complicated Censoring Distribution}
\label{sub:complicated_censoring_distribution}

In the first study, we draw event times using~\eqref{eq:g_all} with two covariates per $\gamma_k$.
The function $Q(t \mid \x)$ is also defined using~\eqref{eq:g_all} and~\eqref{eq:sim_theo_surv}, but with an independent set of covariates, ensuring that the event times are independent of the censoring times.
Note that although the covariates contain enough information to identify the censoring times, the complexity of the censoring distribution makes this a somewhat hard task.

We fit models using all the covariates from both the event-time distribution and the censoring distribution, giving a total of 36 covariates.
We draw 10,000 individuals for training and testing, and 4,000 for a validation set used for early stopping of our training procedure.
The networks are ReLU-nets with 4 layers and 32 nodes in each layer.
Batch normalization and dropout with a rate of 0.1 are applied between each layer.
The BCE and Logistic-Hazard both give estimates for 50 equidistant times points between 0 and 100, 
but we perform constant density interpolation~\citep[linear interpolation of survival estimates, see][]{article3} to obtain predictions for all 1,000 time points.

\begin{figure}[tp]
    \centering
    \includegraphics[width=1.0\linewidth]{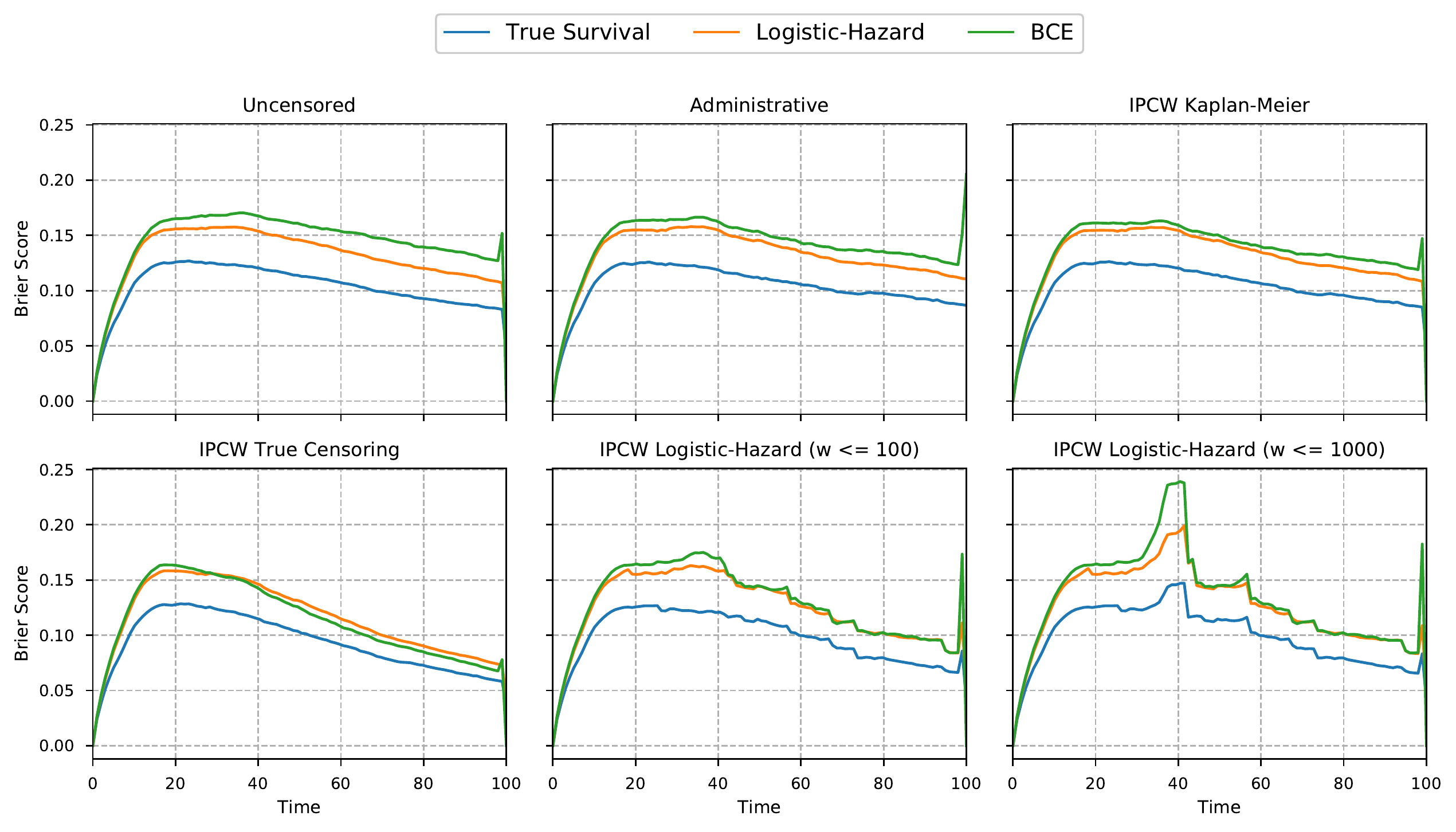}
    \caption{Brier scores from simulations with complicated administrative censoring.}\label{fig:sim_advanced_advanced}
\end{figure}

In Figure~\ref{fig:sim_advanced_advanced}, we have plotted the Brier score of an uncensored test set, the administrative Brier score~\eqref{eq:bs_admin}, and four IPCW Brier scores using~\eqref{eq:bs_ipcw}.
The IPCW scores are calculated with Kaplan-Meier estimates for the censoring distribution, the true censoring function in~\eqref{eq:sim_censor_surv}, and censoring distributions estimated with Logistic-Hazard and  max weights of 100 and 1000.
Recall from Section~\ref{sub:estimating_the_censoring_distribution} that when we estimate the censoring distribution with other methods than Kaplan-Meier, the weights can become very large, resulting in unstable results. Hence we set a max weight, which is given above the respective plots.

From the figure, we see that both methods perform rather poorly compared to the true survival function, and the Logistic-Hazard  performs better than the BCE for all scores except for the IPCW with the true censoring distribution.
The problems with IPCW do not appear here because the functional form of the censoring distribution is quite complicated.
Hence, the BCE method is not able to identify the $\Cs_i$'s to the extent that it can take advantage of the bias of the IPCW scores.

\begin{figure}[tp]
    \centering
    \includegraphics[width=1.0\linewidth]{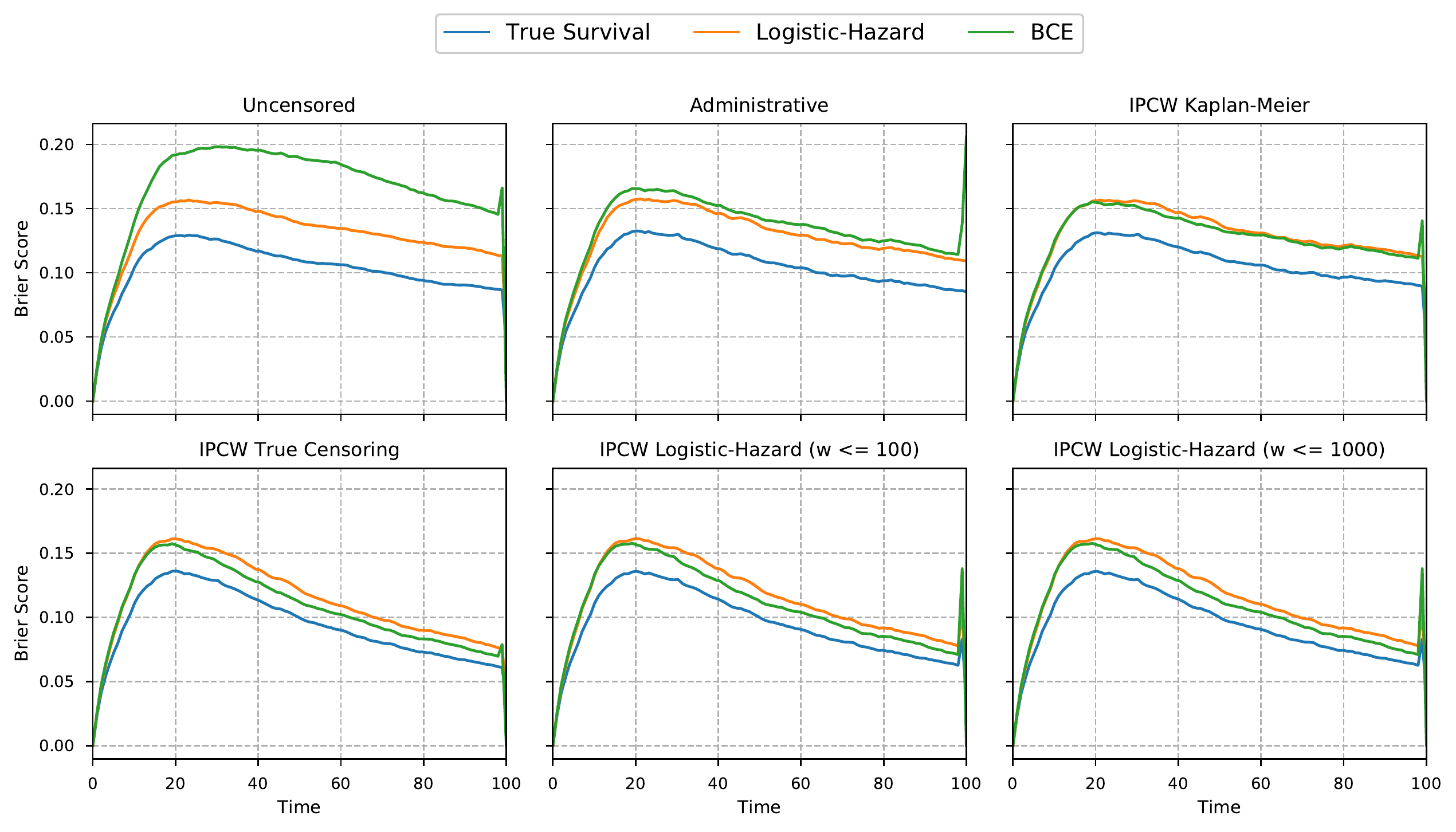}
    \caption{Brier scores from simulations with simple administrative censoring.}\label{fig:sim_advanced_simple}
\end{figure}

\subsection{Simple Censoring Distribution}
\label{sub:simple_censoring_distribution}

In the second simulations study, we let $Q(t \mid \x)$ be defined by~\eqref{eq:sim_theo_surv} with $g(t \mid \x) = \gamma_5$ from~\eqref{eq:g_con} and let $\gamma_5$ be a function of 5 covariates.
This gives a very simple censoring distribution, meaning that the censoring times $\Cs_i$ are quite easily identifiable.
We repeat the simulations with the event-time distribution unchanged, and the results are displayed in Figure~\ref{fig:sim_advanced_simple}.
First, we note that the performance of the BCE method on the uncensored test set is worse than before.
This is expected because it is now easier for the  BCE method to identify the censoring time $\Cs_i$, meaning the survival estimates fall to zero right after $\Cs_i$ (as in Figure~\ref{fig:surv_bce}).
Nevertheless, the BCE method now gets better IPCW scores than the Logistic-Hazard.
This shows that when the administrative censoring function is simple to learn, the BCE method exhibit more of the step behavior of the optimal estimates in~\eqref{eq:bce_optimal_pi}.
We observe, however, that the administrative Brier score rightfully considers the BCE method worse than the Logistic-Hazard.
In this regard, we argue that even though the IPCW Brier score might work well for administrative censoring, the administrative Brier score is the safer choice.

In the simulations, we have considered a censoring distribution with covariates that are independent of the survival covariates, so the event-time distribution is independent of the censoring distribution.
This might explain why the Kaplan-Meier estimates work so well.
If one would encounter this independence for real data, one could simply remove the censoring covariates, as they do not affect the event-time distribution, and use IPCW with Kaplan-Meier weights.

\section{KKBox Churn Prediction}
\label{sec:kkbox_churn_prediction}

Finally, we revisit the KKBox data set discussed in Section~\ref{sec:a_real_world_example}.
We fit the BCE method, corresponding to multiple binary classifiers, and the Logistic-Hazard method.
The training set is of size 100,000 and we use a validation set of size 20,000 for early stopping.
We use 6-layer ReLU networks with 128 nodes in each layer, and with batch normalization and a dropout rate of 0.1 between each layer.
Entity embeddings are used for the categorical covariates \citep{guo2016entity} with embedding sizes that are the square root of the number of categories.
The outputs of the networks are of size 150 representing an equidistant grid of the full time-scale of the training set.
Constant density interpolation~\citep{article3} is applied to the survival estimates to obtain predictions for all the time-points in the test set.
We use the AdamWR~\citep{adamwr} optimizer with a cycle length of 1 epoch, but we double the cycle length and multiply the learning rate by 0.8 after each cycle. Also, we do not include weight decay.
The data set is, essentially, the data set presented by \citet{Cox-Time}, but including all censoring times and an extra categorical covariate stating the payment method.
The code for obtaining the data set is available at \href{https://github.com/havakv/pycox}{\texttt{github.com/havakv/pycox}}.

\begin{figure}[t]
    \centering
    \includegraphics[width=0.99\linewidth]{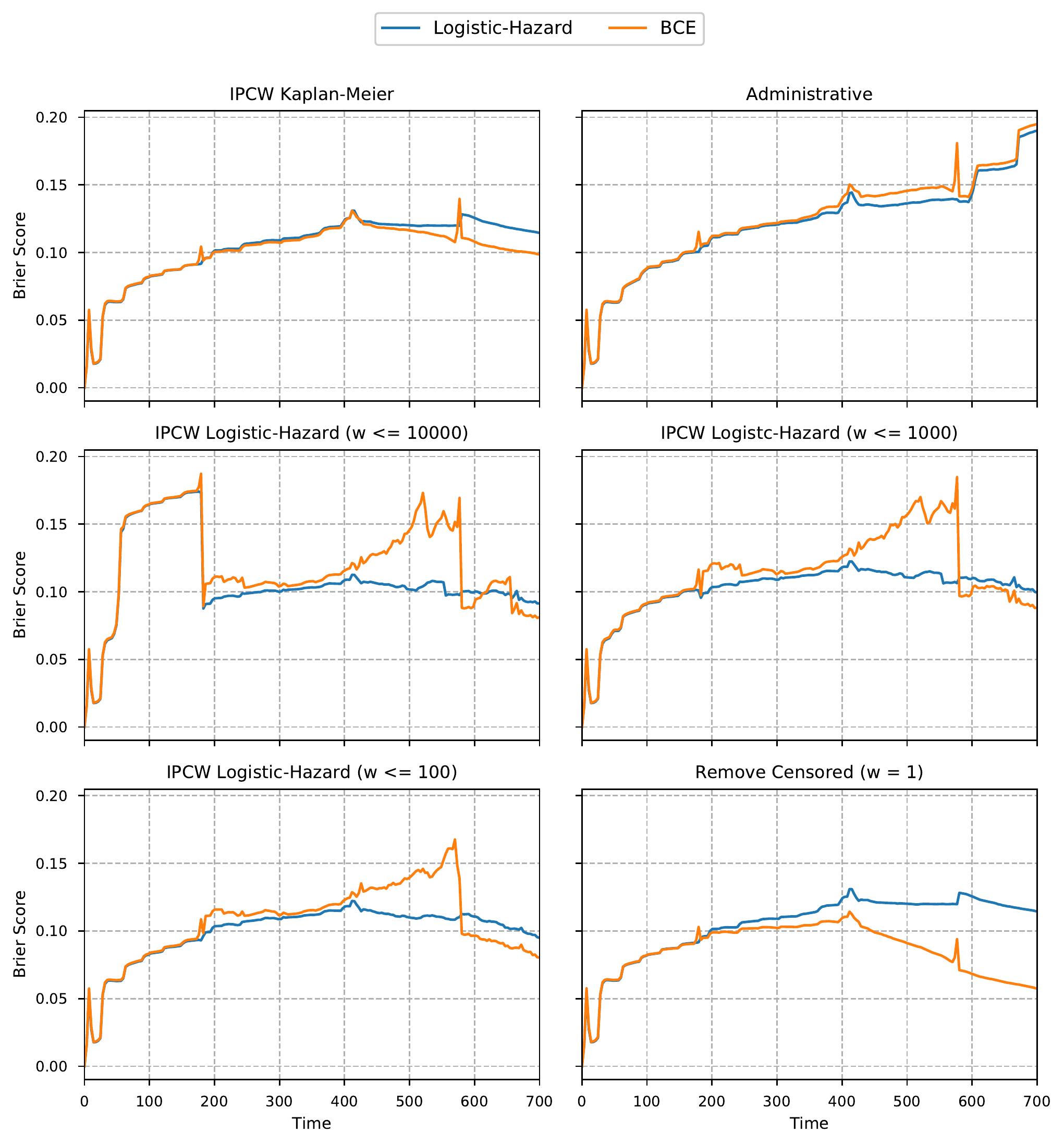}
    \vspace{-0.5cm}
    \caption{Brier scores on the KKBox data set.}\label{fig:kkbox_bs_all}
\end{figure}

The censoring distribution is estimated in two ways. The first is with the  Kaplan-Meier estimator and the second is with a Logistic-Hazard with the same hyperparameters as the Logistic-Hazard used to estimate the churn distribution.
The Brier scores are computed on a test set of size 100,000 and displayed in Figure~\ref{fig:kkbox_bs_all}.
The figure contains the two plots from Figure~\ref{fig:kkbox_bs_km}, the administrative Brier scores~\eqref{eq:bs_admin}, and three other IPCW scores~\eqref{eq:bs_ipcw} with different max weights.
Note that a maximum weight of 1 corresponds to the unweighted score~\eqref{eq:ipcw_admin_n} where the set of censored individuals $\{i:\, T_i \leq t, D_i  =  0\}$ are removed.
This is the same way the BCE method~\eqref{eq:bce_loss_t} handles censored individuals.

As we recall from Section~\ref{sec:a_real_world_example}, the IPCW Brier score with Kaplan-Meier estimates considers the BCE method the best (top left Figure~\ref{fig:kkbox_bs_all}).
We now see from the administrative scores (top right) that this is probably a wrong conclusion, as the BCE method has worse administrative Brier scores than the Logistic-Hazard.
We also see that the IPCW scores obtained with Logistic-Hazard are very dependent on the maximum weight allowed in the scores.
For higher allowed weights, we would consider the BCE method to perform worse than the Logistic-Hazard, but note that for the highest times this is not the case.

The administrative scores are simpler to obtain than the IPCW scores, as they do not require estimation of the censoring distribution. 
Also, from this case study, we see that the administrative scores are probably a safer chose of evaluation metric than the IPCW weighted Brier score.

\section{Discussion}
\label{sec:discussion}

In this paper, we have addressed potential issues of the inverse probability of censoring weighted (IPCW) Brier score, in particular for administrative censoring.
If the covariates have sufficient information to determine the administrative censoring times, the IPCW Brier score will not be minimized by the true survival functions, but instead by a function that falls to zero after the censoring time.
We have also shown that a binary classifier that disregards censored individuals will approach this minimizer.
As a consequence, the binary classifier might actually get better IPCW Brier scores than the true survival function.
Due to this bias, we argue that the IPCW Brier score needs to be applied with care.

The IPCW Brier score can be substantially affected by the estimated censoring distribution.
If this is the case, the validity of the scores can be questioned.
In regards to both issues, we propose the \emph{administrative Brier score} that works for administrative right-censored event times and does not require estimation of the censoring distribution.
This means that it is simpler to use than the IPCW scores.

We simulate examples where the IPCW score fails to identify the best survival estimates, but the administrative Brier score still provides reasonable scores.
We also investigate a real-world churn data set with administrative censoring and find that it exhibits some of the same behavior as our simulations.
This shows that the proposed administrative Brier score can be a very useful evaluation metric.

In this paper, we have only investigated the IPCW Brier score, but note that there are multiple other IPCW scores that might suffer from the same drawbacks as the Brier score.
For example the IPCW Binomial log-likelihood \citep{Graf1999} and the IPCW concordance index \citep{uno2011c,gerds2013estimating}.
Preliminary investigations of the IPCW Binomial log-likelihood \citep{Graf1999} suggest that it has the same issues as the IPCW Brier score while an administrative version of the Binomial log-likelihood behaves in the same manner as the administrative Brier score.

The issues discussed in this paper are mostly relevant for machine learning methods, such as neural networks, as the issues are only notable for quite precise estimates of the event-time distribution and the censoring times $\Cs_i$.
Hence, we are unlikely to encounter such issues with classical statistical models.

Large parts of the machine learning literature rely on empirical evaluation of predictive methodology.
We, therefore, believe that more research on the evaluation metrics for right-censored survival data is needed.



\acks{This work was supported by The Norwegian Research Council 237718 through the Big Insight Center for research-driven innovation.}


\appendix
\setcounter{equation}{0}
\setcounter{table}{0}
\setcounter{figure}{0}
\renewcommand{\theequation}{\thesection.\arabic{equation}}
\renewcommand{\theHequation}{\thesection.\arabic{equation}}
\renewcommand{\thetable}{\thesection.\arabic{table}}
\renewcommand{\theHtable}{\thesection.\arabic{table}}
\renewcommand{\thefigure}{\thesection.\arabic{figure}}
\renewcommand{\theHfigure}{\thesection.\arabic{figure}}
\renewcommand{\theHsection}{\thesection}

\section{The BCE Survival Estimates}
\label{sec:the_bce_survival_estimates}

To better understand the survival estimates of the binary classifiers in Section~\ref{sec:binary_classifiers_for_time_to_event_prediction}, we investigate the minimizers of the expected loss.
Again, we stress that the BCE method corresponds to a set of binary classifiers in the manner given by the loss~\eqref{eq:bce_loss_t}  and~\eqref{eq:bce_loss_full}.

The expected loss of the binary classifier~\eqref{eq:bce_loss_t} is
\begin{align*}
    \mathbb{E}\left[\text{loss}_\text{BCE}(t)\right]
    &= - \sum^{n}_{i=1} \Big( \Pm(T_i > t) \log\left[\pi_i(t)\right]  + \Pm(T_i \leq t,\, D_i = 1) \log \left[1 - \pi_i(t)\right] \Big)\\
    &= - \sum^{n}_{i=1} \Big( \Pm(\Ts_i > t,\, \Cs_i > t) \log\left[\pi_i(t)\right]  + \Pm(\Ts_i \leq t,\, \Cs_i \geq \Ts_i) \log \left[1 - \pi_i(t)\right] \Big)\\
    &= - \sum^{n}_{i=1} \Big( S_i(t) G_i(t) \log\left[\pi_i(t)\right]  +  \left[\int_0^t G_i(u-) f_i(u) du \right] \, \log\left[1 - \pi_i(t)\right] \Big).
\end{align*}
The minimizers of this expectation with respect to $\pi_i(t)$ can be found by equating the partial derivatives with zero,
\begin{align*}
    \fracpartial{\mathbb{E}\left[\text{loss}_\text{BCE}(t)\right]}{\pi_i(t)}
    &= - \frac{S_i(t)\, G_i(t)}{\pi_i(t)} +  \frac{\int_0^t G_i(u-) f_i(u) du}{1 - \pi_i(t)} = 0,
\end{align*}
This gives the minimizers 
\begin{align}
    \label{eq:bce_minimizer}
    \pi_i(t)
    &= \frac{S_i(t) G_i(t)}{S_i(t) G_i(t) + \int_0^t G_i(u-) f_i(u) du}\\
    &\leq \frac{S_i(t) G_i(t)}{S_i(t) G_i(t) + \int_0^t G_i(t) f_i(u) du} \nonumber \\
    &\leq \frac{S_i(t)}{S_i(t) + \int_0^t f_i(u) du} \nonumber \\
    &= S_i(t). \nonumber
\end{align}
We see that the $\pi_i(t)$'s are underestimating the true survival as long as there is censoring present. 
This is expected as the binary classifiers remove censored individuals, decreasing the proportion of survived individuals.

If the censoring times $\Cs_i$ can be identified by the covariates, the censoring distribution is given by the step function
$G_i(t) = \mathbbm{1}\{\Cs_i > t\}$.
The minimizer~\eqref{eq:bce_minimizer} can then be written as
\begin{align*}
    \pi_i(t)
    &= \frac{S_i(t) \mathbbm{1}\{\Cs_i > t\}}{S_i(t) \mathbbm{1}\{\Cs_i > t\} + \int_0^t \mathbbm{1}\{\Cs_i \geq u\} f_i(u) du}.
\end{align*}
If $\Cs_i > t$, we have
\begin{align*}
    \pi_i(t) = \frac{S_i(t)}{S_i(t) + \int_0^t f_i(u) du} = S_i(t),
\end{align*}
and if $\Cs_i \leq t$, we have $\pi_i(t) = 0$.
The minimizer can, therefore, be written as
\begin{align*}
    \pi_i(t) = S_i(t) \mathbbm{1}\{ \Cs_i > t\}.
\end{align*}
We recognize this as the minimizer of the IPCW Brier score with administrative censoring from Section~\ref{sub:ipcw_with_administrative_censoring}.
So if there is sufficient information in the covariates to identify the administrative censoring times $\Cs_i$, the binary classifiers will approach the minimizers of the IPCW Brier score.



\vskip 0.2in
\bibliography{bibliography}

\end{document}